\documentclass[11pt,a4paper]{article}

\pdfoutput=1

\usepackage[nohyperref,acceptedwitha]{tacl2018v2}
\usepackage[font=small]{caption}
\usepackage{natbib}
\usepackage[hyphens]{url}
\usepackage{collcell}
\usepackage{graphicx}
\usepackage[utf8]{inputenc}
\usepackage{latexsym}
\usepackage{linguex}
\usepackage{multirow}
\usepackage{pgfplots}
\usepackage{slashbox}
\usepackage{tikz}
\usepackage{times}
\usepackage{url}
\usepackage{xstring}
\usepackage{tabulary}

\usepackage{enumitem}
\setitemize{noitemsep,topsep=0pt,parsep=0pt,partopsep=0pt}

\pgfplotsset{compat=newest}

\newcommand{\todo}[1]{}
\renewcommand{\todo}[1]{{\color{red} TODO: {#1}}}

\newcommand*{\MinNumber}{42}
\newcommand*{\MaxNumber}{60}
\newcommand{\ApplyGradient}[1]{%
    \pgfmathsetmacro{\PercentColor}{100.0*(#1-\MinNumber)/(\MaxNumber-\MinNumber)}
    \colorbox{blue!\PercentColor!white}{#1}
}
\newcolumntype{R}{>{\collectcell\ApplyGradient}p{0.85cm}<{\endcollectcell}}

\title{Mind the GAP: \\ A Balanced Corpus of Gendered Ambiguous Pronouns}
\author{Kellie Webster \and  Marta Recasens \and Vera Axelrod \and Jason Baldridge \\ Google AI Language \\ {\tt \{websterk|recasens|vaxelrod|jasonbaldridge\}@google.com}}
\date{}

\begin{document}

\maketitle

\begin{abstract}
Coreference resolution is an important task for natural language understanding, and the resolution of ambiguous pronouns a longstanding challenge.
Nonetheless, existing corpora do not capture ambiguous pronouns in sufficient volume or diversity to accurately indicate the practical utility of models.
Furthermore, we find gender bias in existing corpora and systems favoring masculine entities.
To address this, we present and release GAP, a gender-balanced labeled corpus of 8,908 ambiguous pronoun-name pairs sampled to provide diverse coverage of challenges posed by real-world text.
We explore a range of baselines which demonstrate the complexity of the challenge, the best achieving just 66.9\% F1. 
We show that syntactic structure and continuous neural models provide promising, complementary cues for approaching the challenge.
\end{abstract}

\section{Introduction}

Coreference resolution involves linking referring expressions that evoke the same discourse entity, as defined in shared tasks such as CoNLL 2011/12 \citep{pradhan2012} and MUC \citep{grishman1996}. Unfortunately, high scores on these tasks do not necessarily translate into acceptable performance for downstream applications such as machine translation \citep{guillou2012} and fact extraction \citep{nakayama2008}. 
In particular, high-scoring systems successfully identify coreference relationships between string-matching proper names, but fare worse on anaphoric mentions such as pronouns and common noun phrases \citep{stoyanov2009,rahman2012,DurrettKlein2013}. 

\noindent We consider the problem of resolving gendered ambiguous pronouns in English, such as \textbf{she}\footnote{The examples throughout the paper highlight the ambiguous pronoun in bold, the two potential coreferent names in italics, and the correct one also underlined.} in:

\ex. \footnotesize \label{ex:intro} In May, \underline{\emph{Fujisawa}} joined \emph{Mari Motohashi}'s rink as the team's skip, moving back from Karuizawa to Kitami where \textbf{she} had spent her junior days.

With this scope, we make three key contributions:

\begin{itemize}
\item We design an extensible, language-independent mechanism for extracting challenging ambiguous pronouns from text.
\item We build and release GAP, a human-labeled corpus of 8,908 ambiguous pronoun-name pairs derived from Wikipedia.\footnote{\url{http://goo.gl/language/gap-coreference}}
This dataset targets the challenges of resolving naturally-occurring ambiguous pronouns and rewards systems which are gender-fair.
\item We run four state-of-the-art coreference resolvers and several competitive simple baselines on GAP to understand limitations in current modeling, including gender bias. We find that syntactic structure and Transformer models \citep{vaswani:etal:2017} provide promising, complementary cues for approaching GAP.
\end{itemize}

\noindent
Coreference resolution decisions can drastically alter how automatic systems process text.
Biases in automatic systems have caused a wide range of underrepresented groups to be served in an inequitable way by downstream applications \citep{hardt2014}.
We take the construction of the new GAP corpus as an opportunity to reduce gender bias in coreference datasets;
in this way, GAP can promote equitable modeling of reference phenomena complementary to the recent work of \citet{zhao2018} and \citet{rudinger2018}.
Such approaches promise to improve equity of downstream models, such as triple extraction for knowledge base population.

\section{Background}

Existing datasets do not capture ambiguous pronouns in sufficient volume or diversity to benchmark systems for practical applications. 

\subsection{Datasets with Ambiguous Pronouns}

Winograd schemas \citep{levesque2012} are closely related to our work as they contain ambiguous pronouns. They are pairs of short texts with an ambiguous pronoun and a special word (in square brackets) that switches its referent:

\ex. \footnotesize \label{ex:winograd} \emph{The trophy} would not fit in \emph{the brown suitcase} because \textbf{it} was too [big/small].

\noindent
The Definite Pronoun Resolution Dataset \citep{rahman2012} comprises 943 Winograd schemas written by undergraduate students and later extended by \citet{peng2015}. 
The First Winograd Schema Challenge \citep{morgenstern2016} released 60 examples adapted from published literary works (Pronoun Disambiguation Problem)\footnote{\url{https://cs.nyu.edu/faculty/davise/papers/WinogradSchemas/PDPChallenge2016.xml}} and 285 manually constructed schemas (Winograd Schema Challenge)\footnote{\url{https://cs.nyu.edu/faculty/davise/papers/WinogradSchemas/WSCollection.xml}}.
More recently, \citet{rudinger2018} and \citet{zhao2018} have created two Winograd schema-style datasets containing 720 and 3160 sentences, respectively, where each sentence contains a gendered pronoun and two occupation (or participant) antecedent candidates that break occupational gender stereotypes.
Overall, ambiguous pronoun datasets have been limited in size and, most notably, consist only of manually constructed examples which do not necessarily reflect the challenges faced by systems in the wild.

In contrast, the largest and most widely-used coreference corpus, OntoNotes \citep{pradhan2007}, is general purpose. In OntoNotes, simpler high-frequency coreference examples (e.g. those captured by string matching) greatly outnumber examples of ambiguous pronouns, which obscures performance results on that key class \citep{stoyanov2009,rahman2012}. 
Ambiguous pronouns greatly impact main entity resolution in Wikipedia, the focus of \citet{ghaddar2016}, who use WikiCoref, a corpus of 30 full articles annotated with coreference \citep{ghaddar2016lrec}.

GAP examples are not strictly Winograd schemas because they have no reference-flipping word. Nonetheless, they contain two person named entities of the same gender and an ambiguous pronoun that may refer to either (or neither). As such, they represent a similarly difficult challenge and require the same inferential capabilities. More importantly, GAP is larger than existing Winograd schema datasets and the examples are from naturally occurring Wikipedia text. GAP complements OntoNotes by providing an extensive targeted dataset of naturally occurring ambiguous pronouns.

\begin{table*}[]
    \centering
    {\small
    \begin{tabular}{ccp{9cm}}
        Type & Pattern & Example \\
        \hline
        \textsc{FinalPro}   & (Name, Name, Pronoun)      
        & \emph{Preckwinkle} criticizes \underline{\emph{Berrios'}} nepotism: [\dots] County's ethics rules don't apply to \textbf{him}. \\
        \hline
        \textsc{MedialPro}  & (Name, Pronoun, Name)      
        & \underline{\emph{McFerran}}'s horse farm was named Glen View. After \textbf{his} death in 1885, \emph{John E. Green} acquired the farm. \\
        \hline
        \textsc{InitialPro} & (Pronoun, Name, Name) 
        & Judging that \textbf{he} is suitable to join the team, \emph{Butcher} injects \underline{\emph{Hughie}} with a specially formulated mix.
    \end{tabular}
    }
    \caption{Extraction patterns and example contexts for each.}
    \label{tab:patterns}
\end{table*}

\subsection{Modeling Ambiguous Pronouns}

State-of-the-art coreference systems struggle to resolve ambiguous pronouns that require world knowledge and commonsense reasoning \citep{DurrettKlein2013}. Past efforts have tried to mine semantic preferences and inferential knowledge via predicate-argument statistics mined from corpora \citep{dagan1990,yang2005}, semantic roles \citep{kehler2004,ponzetto2006}, contextual compatibility features \citep{liao2010,bansal2012}, and event role sequences \citep{bean2004,chambers2008}. These usually bring small improvements in general coreference datasets and larger improvements in targeted Winograd datasets.

\citet{rahman2012} scored 73.05\% precision on their Winograd dataset after incorporating targeted features such as narrative chains, Web-based counts, and selectional preferences.
\citet{peng2015}'s system improved the state of the art to 76.41\% by acquiring $\langle$subject, verb, object$\rangle$ and $\langle$subject/object, verb, verb$\rangle$ knowledge triples.

In the First Winograd Schema Challenge \citep{morgenstern2016}, participants used methods ranging from logical axioms and inference to neural network architectures enhanced with commonsense knowledge \citep{liu2017}, but no system qualified for the second round. Recently, \citet{trinh2018} have achieved the best results on the Pronoun Disambiguation Problem and Winograd Schema Challenge datasets, achieving 70\% and 63.7\%, respectively, which are 3\% and 11\% above \citeauthor{liu2017}'s (\citeyear{liu2017})'s previous state of the art. Their model is an ensemble of word-level and character-level recurrent language models, which despite not being trained on coreference data, encode commonsense as part of the more general language modeling task. It is unclear how these systems perform on naturally-occurring ambiguous pronouns. For example, \citeauthor{trinh2018}'s (\citeyear{trinh2018}) system relies on choosing a candidate from a pre-specified list, and it would need to be extended to handle the case that the pronoun does not corefer with any given candidate. By releasing GAP, we aim to foster research in this direction, and set several competitive baselines without using targeted resources.

\subsection{Bias in Machine Learning}
\label{sec:bias-datas}

While existing corpora have promoted research into coreference resolution, they suffer from gender bias.
Specifically, of the over 2,000 gendered pronouns in the OntoNotes test corpus, less than 25\% are feminine \citep{zhao2018}.
The imbalance is more pronounced on the development and training sets, with less than 20\% feminine pronouns each. 
WikiCoref contains only 12\% feminine pronouns. 
In the Definite Pronoun Resolution Dataset training data, 27\% of the gendered pronouns are feminine, while the Winograd Schema Challenge datasets contain 28\% and 33\% feminine examples.
Two exceptions are the recent WinoBias \citep{zhao2018} and Winogender schemas \citep{rudinger2018} datasets, which reveal how occupation-specific gender bias pervades in the majority of publicly-available coreference resolution systems by including a balanced number of feminine pronouns that corefer with anti-stereotypical occupations (see~Example~\ref{ex:winobias} from WinoBias). These datasets focus on pronominal coreference where the antecedent is a nominal mention, while GAP focuses on relations where the antecedent is a named entity.

\ex. \footnotesize \label{ex:winobias} \underline{\emph{The salesperson}} sold some books to \emph{the librarian} because \textbf{she} was trying the sell them.

The pervasive bias in existing datasets is concerning given that learned NLP systems often reflect and even amplify training biases \citep{bolukbasi2016,caliskan2017,zhao2017}.
A growing body of work defines notions of fairness, bias, and equality in data and machine-learned systems \citep{pedreshi2008, hardt2016, zafar2017, skirpan2017}, and debiasing strategies include expanding and rebalancing data \citep{torralba2011,ryu2017,shankar2017,buda2017}, and balancing performance across subgroups \citep{dwork2012}.
In the context of coreference resolution, \citet{zhao2018} have showed how debiasing tecniques (e.g. swapping the gender of male pronouns and antecedents in OntoNotes, using debiased word embeddings, balancing \citeauthor{bergsma2006}'s (\citeyear{bergsma2006})'s gender list) succeed at reducing the gender bias of multiple off-the-shelf coreference systems.

We work towards fairness in coreference by releasing a diverse, gender-balanced corpus for ambiguous pronoun resolution and further investigating performance differences by gender, not specifically on pronouns with an occupation antecedent but more generally on gendered pronouns.

\section{GAP Corpus}

\begin{table*}[]
    \centering  \small
    \begin{tabular}{l|l|r}
        Dimension           & Values                    & Ratio          \\
        \hline
        Page coverage       &                           & 1 per page per \\
                            &                           & pronoun form   \\
        Gender              & masc. : fem.              & 1 : 1          \\
        Extraction Pattern  & final : medial : initial  & 6.2 : 1 : 1    \\
        Page Entity         & true : false              & 1.3 : 1        \\
        Coreferent Name     & nameA : nameB             & 1 : 1          \\
    \end{tabular}
    \caption{Corpus diversity statistics in final corpus.}
    \label{tab:diversity stats}
\end{table*}

We create a corpus of 8,908-human annotated ambiguous pronoun-name examples from Wikipedia. 
Examples are obtained from a large set of candidate contexts and are filtered through a multi-stage process designed to improve quality and diversity. 

We choose Wikipedia as our base dataset given its wide use in natural language understanding tools, but are mindful of its well-known gender biases.
Specifically, less than 15\% of biographical Wikipedia pages are about women. 
Furthermore, women are written about differently than men:
e.g. women's biographies are more likely to mention marriage or divorce \citep{bamman2014}, 
 abstract terms are more positive in male biographies than female biographies \citep{wagner2016},
and articles about females are less central to the article graph \citep{graells2015}.

\subsection{Extraction and Filtering}

Extraction targets three patterns, given in Table~\ref{tab:patterns}, that characterize locally ambiguous pronoun contexts. 
We limit to singular mentions, gendered non-reflexive pronouns, and names whose head tokens are different from one another.
Additionally, we do not allow intruders: 
there can be no other compatible mention (by gender, number, and entity type) between the pronoun and the two names.

To limit the success of na\"ive resolution heuristics, we apply a small set of constraints to focus on those pronouns that are truly hard to resolve.

\begin{itemize}
\item \textbf{\textsc{FinalPro}}. Both names must be in the same sentence, 
and the pronoun may appear in the same or directly following sentence.
\item \textbf{\textsc{MedialPro}}. The first name must be in the sentence directly preceding the pronoun and the second name, both of which are in the same sentence. 
To decrease the bias for the pronoun to be coreferential with the first name, the pronoun must be in an initial subordinate clause or be a possessive in an initial prepositional phrase.
\item \textbf{\textsc{InitialPro}}. All three mentions must be in the same sentence and the pronoun must be in an initial subordinate clause or a possessive in an initial prepositional phrase.
\end{itemize}

From the extracted contexts, we sub-sample those to send for annotation.
We do this to improve diversity in five dimensions:

\begin{itemize}
\item \textbf{Page Coverage}. We retain at most 3 examples per page-gender pair to ensure a broad coverage of domains.
\item \textbf{Gender}. The raw pipeline extracts contexts with a m:f ratio of 9:1. We oversampled feminine pronouns to achieve a 1:1 ratio.\footnote{In doing this, we observed that many feminine pronouns in Wikipedia refer to characters in film and television.}
\item \textbf{Extraction Pattern}. The raw pipeline output contains 7 times more \textsc{FinalPro} contexts than \textsc{MedialPro} and \textsc{InitialPro} combined, so we oversampled the latter two to lower the ratio to 6:1:1.
\item \textbf{Page Entity}. Pronouns in a Wikipedia page often refer to the entity the page is about. We include such examples in our dataset but balance them 1:1 against examples that do not include mentions of the page entity.
\item \textbf{Coreferent Name}. To ensure mention order is not a cue for systems, our final dataset is balanced for label --- i.e. whether Name A or Name B is the pronoun's referent. 
\end{itemize}

We applied these constraints to the raw extractions to select 8,604 contexts (17,208 examples) for annotation that were globally balanced in all dimensions (e.g. $~$1:1 gender ratio in \textsc{MedialPro} extractions).
Table~\ref{tab:diversity stats} summarizes the diversity ratios obtained in the final dataset, whose compilation is described next.

\subsection{Annotation}

We used a pool of in-house raters for human annotation of our examples.
Each example was presented to three workers, who selected one of five labels (Table~\ref{tab:label-stats}). 
Full sentences of at least 50 tokens preceding each example were presented as context (prior context beyond a section break is not included).
Rating instructions accompany the dataset release. 

Despite workers not being expert linguists, we find good agreement both within workers and between workers and an expert.
Inter-annotator agreement was $\kappa$ = 0.74 on the \citet{fleiss} Kappa statistic;
in 73\% of cases there was full agreement between workers, in 25\% of cases two of three workers agreed, and only in 2\% of cases there was no consensus.
We discard the 194 cases with no consensus.
On 30 examples rated by an expert linguist, there was agreement on 28 and one was deemed to be truly ambiguous with the given context.

To produce our final dataset, we applied additional high-precision filtering to remove some error cases identified by workers,\footnote{E.g. missing sentence breaks, list environments, and non-referential personal roles/nationalities.} and discarded the ``Both" (no ambiguity) and ``Not Sure" contexts. Given that many of the feminine examples received the ``Both" label from referents having stage and married names \ref{femex}, this unbalanced the number of masculine and feminine examples.
 
\ex. \footnotesize \label{femex} \emph{Ruby Buckton} is a fictional character from the Australian Channel Seven soap opera Home and Away, played by \emph{Rebecca Breeds}. \textbf{She} debuted  \dots

To correct this, we discarded masculine examples to re-achieve 1:1 gender balance.
Additionally, we imposed the constraint that there be one example per Wikipedia article per pronoun form (e.g. \textbf{his}), to reduce similarity between examples.
The final counts for each label are given in the second column of Table~\ref{tab:label-stats}.
Given that the 4,454 contexts each contain two annotated names, this comprises 8,908 pronoun-name pair labels.

\begin{table}[]
    \centering \small
    \begin{tabular}{l|rr}
        Label                       & Raw  & Final \\
        \hline
        Name A                      & 2913 & 1979 \\  
        Name B                      & 3047 & 1985 \\ 
        Neither Name A nor Name B   & 1614 &  490 \\ 
        Both Name A and Name B      & 1016 &    0 \\ 
        Not Sure                    &   14 &    0 \\
        \hline
        {\bf Total}                 & 8604 & 4454 \\ 
    \end{tabular}
    \caption{Consensus label counts for the extracted examples (Raw) and after further filtering (Final).}
    \label{tab:label-stats}
\end{table}

\section{Experiments}

We set up the GAP challenge and analyze the applicability of a range of off-the-shelf tools.
We find that existing resolvers do not perform well and are biased to favor better resolution of masculine pronouns.
We empirically validate the observation that Transformer models \citep{vaswani:etal:2017} encode coreference relationships, adding to the results by \citet{voita2018transformers} on machine translation, and \citet{trinh2018} on language modeling.
Furthermore, we show they complement traditional linguistic cues such as syntactic distance and parallelism.

All experiments use the Google Cloud NL API\footnote{\url{https://cloud.google.com/natural-language/}} for pre-processing, unless otherwise noted.

\subsection{GAP Challenge}

GAP is an evaluation corpus and we segment the final dataset into a development and test set of 4,000 examples each\footnote{All examples extracted from the same URL are partitioned into the same set.};
we reserve the remaining 908 examples as a small validation set for parameter tuning.
All examples are presented with the URL of the source Wikipedia page, allowing us to define two task settings: \textit{snippet-context} in which the URL may not be used, and \textit{page-context} in which it may.
While name spans are given in the data, we urge the community not to treat this as a \textit{gold mention} or Winograd-style task.
That is, systems should detect mentions for inference automatically, and access labeled spans only to output predictions.

To reward unbiased modeling, we define two evaluation metrics: F1 score and Bias.
Concretely, we calculate F1 score \textbf{O}verall as well as by the gender of the pronoun (\textbf{M}asculine and \textbf{F}eminine).
\textbf{B}ias is calculated by taking the ratio of feminine to masculine F1 scores, typically less than one.\footnote{\url{http://goo.gl/language/gap-coreference}}

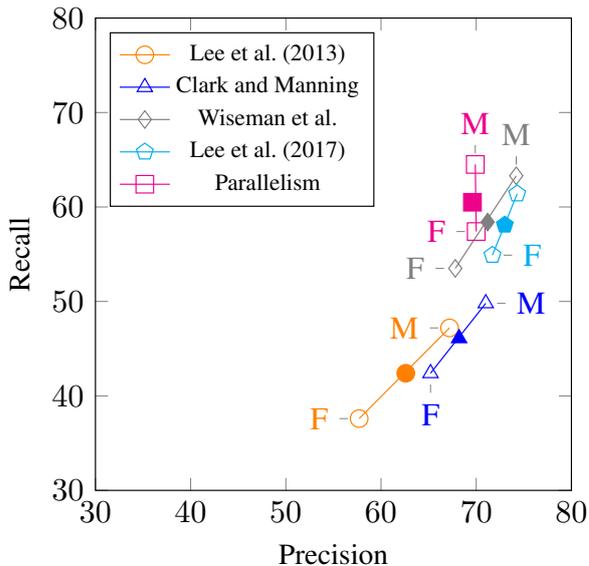
\begin{figure}
    \centering
    \begin{tikzpicture}[scale=1.10, pin distance=1mm]
        \begin{axis}[xlabel=Precision, ylabel=Recall, legend pos=north west,
                     xmin=30, xmax=80, ymin=30, ymax=80, mark size=3,
                     xlabel style={font=\small},
                     ylabel style={font=\small},
                     legend style={font=\scriptsize},
                     axis equal image
                     ]
        \addplot[mark=o, color=orange] coordinates { 
            (67.2,  47.2)  
            (57.7,  37.6)  
        }
        node[pos=1.0, pin=left:F]{}
        node[pos=0.0, pin=left:M]{}
        ;
        \addlegendentry{\citet{heeyoung2013}}
        
        \addplot[mark=triangle, color=blue] coordinates {
            (71.0,  49.8)  
            (65.2,  42.4)  
        }
        node[pos=1.0, pin=south:F]{}
        node[pos=0.0, pin=right:M]{}
        ;
        \addlegendentry{\citeauthor{clark2015}}
        
        \addplot[mark=diamond, color=gray] coordinates {
            (74.2,  63.3)  
            (67.8,  53.5)  
        }
        node[pos=1.0, pin=left:F]{}
        node[pos=0.0, pin=north:M]{}
        ;
        \addlegendentry{\citeauthor{wiseman2016}}
        
        \addplot[mark=pentagon, color=cyan] coordinates {
            (74.3,  61.4)  
            (71.7,  54.9)  
        }
        node[pos=1.0, pin=right:F]{}
        ;
        \addlegendentry{\citet{lee2017}}
        
        \addplot[mark=square, color=magenta] coordinates {
            (69.9,  64.5)  
            (70.0,  57.4)  
        }
        node[pos=1.0, pin=left:F]{}
        node[pos=0.0, pin=north:M]{}
        ;
        \addlegendentry{Parallelism}
        
        \addplot[mark=*, color=orange] coordinates {
            (62.6,  42.4)  
        };
        \addplot[mark=triangle*, color=blue] coordinates {
            (68.2,  46.1)  
        };
        \addplot[mark=diamond*, color=gray] coordinates {
            (71.2, 58.4)  
        };
        \addplot[mark=pentagon*, color=cyan] coordinates {
            (73.0,  58.1)  
        };
        \addplot[mark=square*, color=magenta] coordinates {
            (69.6,  60.5)  
        };
        \end{axis}
    \end{tikzpicture}
    \caption{Precision-Recall on the GAP development dataset---Overall (solid markers), \textbf{M}asculine, \textbf{F}eminine---for off-the-shelf resolvers and Parallelism.}
    \label{fig:pr-plot}
\end{figure}

\subsection{Off-the-Shelf Resolvers}
\label{sec:ontonotes-expts}

The first set of baselines we explore are four representative off-the-shelf coreference systems:
the rule-based system of \citet{heeyoung2013} and three neural resolvers---\citet{clark2015}\footnote{\url{https://stanfordnlp.github.io/CoreNLP/download.html}}, \citet{wiseman2016}\footnote{\url{https://github.com/swiseman/nn_coref}}, and \citet{lee2017}\footnote{\url{https://github.com/kentonl/e2e-coref}}.
All were trained on OntoNotes and run in as close to their out-of-the-box configuration as possible.\footnote{
We run \citet{lee2017} in the \texttt{final} (single-model) configuration, with NLTK preprocessing \citep{nltk}; for
\citet{wiseman2016} we use Berkeley preprocessing \citep{DurrettKlein2014} and the Stanford systems are run within Stanford CoreNLP \citep{manning2014}.
}
System clusters were scored against GAP examples according to whether the cluster containing the target pronoun also contained the correct name~(\textbf{TP}) or the incorrect name~(\textbf{FP}), using mention heads for alignment.
We report here their performance on GAP as informative baselines, but expect retraining on Wikipedia-like texts to yield an overall improvement in performance. (This remains as future work.)

Table~\ref{tab:ontonotes-baselines} shows that all systems struggle on GAP.
That is, despite modeling improvements in recent years, ambiguous pronoun resolution remains a challenge.
We note particularly the large difference in performance between genders, which traditionally has not been tracked but has fairness implications for downstream tasks using these publicly available models.

\begin{table}[]
    \centering
    \begin{tabular}{l|rrrr}
    & \multicolumn{1}{c}{M} & \multicolumn{1}{c}{F} & \multicolumn{1}{c}{B} & \multicolumn{1}{c}{O}     \\
    \hline
    \citet{heeyoung2013}       & 55.4          & 45.5          & \textit{0.82}          & 50.5          \\
    \citeauthor{clark2015}     & 58.5          & 51.3          & \textit{0.88}          & 55.0          \\
    \citeauthor{wiseman2016}   & \textbf{68.4} & 59.9          & \textit{0.88}          & 64.2          \\
    \citet{lee2017}            & 67.2          & \textbf{62.2} & \textbf{\textit{0.92}} & \textbf{64.7} \\
    \end{tabular}
    \caption{Performance of off-the-shelf resolvers on the GAP development set, split by \textbf{M}asculine and \textbf{F}eminine (\textbf{B}ias shows F/M), and \textbf{O}verall. Bold indicates best performance.}
    \label{tab:ontonotes-baselines}
\end{table}

Table~\ref{tab:ontonotes-gender} provides evidence that this low performance is not solely due to domain and task differences between GAP and OntoNotes.
Specifically, with the exception of \citet{clark2015}, the table shows that system performance on pronoun-name coreference relations in the OntoNotes test set\footnote{
For each gendered pronoun in a gold OntoNotes cluster, we compare the system cluster with that pronoun. 
We count a \textbf{TP} if the system entity contains at least one gold coreferent NE mention; \textbf{FP} if the system entity contains at least one non-gold NE mention, and \textbf{FN} if the system entity does not contain any gold NE mention.} is not vastly better compared to GAP.
One possible reason that in-domain OntoNotes performance and out-of-domain GAP performance are not very different could be that state-of-the-art systems are highly tuned for resolving names rather than ambiguous pronouns.

Further, the relative performance of the four systems is different on GAP than on OntoNotes.
Particularly interesting is that the current strongest system overall for OntoNotes, namely \citet{lee2017}, scores best on GAP pronouns but has the largest gender bias on OntoNotes.
This perhaps is not surprising given the dominance of masculine examples in that corpus.
It is outside the scope of this paper to provide an in-depth analysis of the data and modeling decisions which cause this bias; instead we release GAP to address the measurement problem behind the bias.

\begin{table}[]
    \centering
    \begin{tabular}{l|rrrr}
    & \multicolumn{1}{c}{M} & \multicolumn{1}{c}{F} & \multicolumn{1}{c}{B} & \multicolumn{1}{c}{O} \\
    \hline
    \citet{heeyoung2013}     & 47.7          & 53.2          & \textit{1.12}          & 49.2 \\
    \citeauthor{clark2015}   & 64.3          & \textbf{63.9} & \textbf{\textit{0.99}} & \textbf{64.2} \\ 
    \citeauthor{wiseman2016} & 61.9          & 58.0          & \textit{0.94}          & 60.6 \\
    \citet{lee2017}          & \textbf{68.9} & 51.9          & \textit{0.75}          & 63.4 \\
    \end{tabular}
    \caption{Pronoun-name F1 score, by gender, of off-the-shelf systems on the OntoNotes test set. Scores based on 2091 masculine pronoun-named entity pairs (in 403 clusters) and 1095 feminine pairs (in 104 clusters). Bold indicates best performance.}
    \label{tab:ontonotes-gender}
\end{table}

Figure~\ref{fig:pr-plot} compares the recall/precision trade-off for each system split by Masculine and Feminine examples, as well as combined (Overall).
Also shown is a simple syntactic Parallelism heuristic in which subject and direct object pronoun are resolved to names with the same grammatical role (see Section~\ref{sec:coreference-cue-baselines}).
In this visualization, we see a further factor contributing to the low performance of off-the-shelf systems, namely their low recall.
That is, while personal pronouns are overwhelmingly anaphoric in both OntoNotes and Wikipedia texts, OntoNotes-trained models are conservative. 
This observation is consistent with the results for \citet{heeyoung2013} on the Definite Pronoun Resolution Dataset \citep{rahman2012}, on which the system scored 47.2\% F1,\footnote{Calculated based on the reported performance of 40.07\% Correct, 29.79\% Incorrect, and 30.14\% No decision.} failing to beat a random baseline due to conservativeness.

\begin{table}[t]
    \centering
    \begin{tabular}{l|rrrr}
    & \multicolumn{1}{c}{M} & \multicolumn{1}{c}{F} & \multicolumn{1}{c}{B} & \multicolumn{1}{c}{O} \\
    \hline
    Random               & 43.6 & 39.3 & \textit{0.90} & 41.5 \\
    Token Distance       & 50.1 & 42.4 & \textit{0.85} & 46.4 \\
    Topical Entity       & 51.5 & 43.7 & \textit{0.85} & 47.7 \\
    \hline
    Syntactic Distance   & 63.0 & 56.2 & \textit{0.89} & 59.7 \\
    Parallelism          & \textbf{67.1} & \textbf{63.1} & \textbf{\textit{0.94}} & \textbf{65.2} \\
    \hline
    Parallelism+URL      & \textbf{71.1} & \textbf{66.9} & \textbf{\textit{0.94}} & \textbf{69.0} \\
    \hline
    Transformer-Single   & 58.6 & 51.2 & \textit{0.87} & 55.0 \\
    Transformer-Multi    & 59.3 & 52.9 & \textit{0.89} & 56.2 \\
    \end{tabular}
    \caption{Performance of our baselines on the development set. Parallelism+URL tests the page-context setting; all other test the snippet-context setting. Bold indicates best performance in each setting.}
    \label{tab:baselines}
\end{table}

\subsection{Coreference-Cue Baselines}
\label{sec:coreference-cue-baselines}

To understand the shortcomings of state-of-the-art coreference systems on GAP, the upper sections of Table~\ref{tab:baselines} consider several simple baselines based on traditional cues for coreference.

To calculate these baselines, we first detect candidate antecedents by finding all mentions of \textsc{person} entity type, \textsc{name} mention type (headed by a proper noun), and, for structural cues, that are not in a syntactic position which precludes coreference with the pronoun. 
We do not require gender match because gender annotations are not provided by the Google Cloud NL API and, even if they were, gender predictions on last names (without the first name) are not reliable in the \textit{snippet-context} setting.
Second, we select among the candidates using one of the heuristics described next.

For scoring purposes, we do not require exact string match for mention alignment, that is, if the selected candidate is a substring of a given name (or vice versa), we infer a coreference relation between that name and the target pronoun.\footnote{
Note that requiring exact string match drops recall and causes only a small difference in F1 performance.}

\paragraph{Surface Cues}

Baseline cues which require only access to the input text are:

\begin{itemize}
    \item \textbf{\textsc{Random}}. Select a candidate uniformly at random.
    \item \textbf{\textsc{Token Distance}}. Select the closest candidate to the pronoun, with distance measured as the number of tokens between spans. 
    \item \textbf{\textsc{Topical Entity}}. Select the closest candidate which contains the most frequent token string among extracted candidates.
\end{itemize}

\noindent The performance of \textsc{Random} (41.5 Overall) is lower than an otherwise possible guess rate of $\sim$50\%.
This is because the baseline considers all possible candidates, not just the two annotated names. 
Moreover, the difference between masculine and feminine examples suggests that there are more distractor mentions in the context of feminine pronouns in GAP.
To measure the impact of pronoun context, we include performance on the artificial \textit{gold-two-mention} setting where only the two name spans are candidates for inference (Table~\ref{tab:baselines-2cands}).
\textsc{Random} is indeed closer here to the expected 50\% and other baselines are closer to gender-parity.

\textsc{Token Distance} and \textsc{Topical Entity} are only weak improvements above \textsc{Random}, validating that our dataset creation methodology controlled for these factors.

\begin{table}[]
    \centering
    \begin{tabular}{l|rrrr}
    & \multicolumn{1}{c}{M} & \multicolumn{1}{c}{F} & \multicolumn{1}{c}{B} & \multicolumn{1}{c}{O} \\
    \hline
    Random               & 47.5 & 50.5 & \textit{1.06} & 49.0 \\
    Token Distance       & 50.6 & 47.5 & \textit{0.94} & 49.1 \\
    Topical Entity       & 50.2 & 47.3 & \textit{0.94} & 48.8 \\
    \hline
    Syntactic Distance   & 66.7 & 66.7 & \textbf{\textit{1.00}} & 66.7 \\
    Parallelism          & \textbf{69.3} & \textbf{69.2} & \textbf{\textit{1.00}} & \textbf{69.2} \\
    \hline
    Parallelism+URL      &  \textbf{74.2}  & \textbf{71.6} & \textbf{\textit{0.96}} & \textbf{72.9} \\
    \hline
    Transformer-Single   & 59.6 & 56.6 & \textit{0.95} & 58.1 \\
    Transformer-Multi    & 62.9 & 61.7 & \textit{0.98} & 62.3 \\
    \end{tabular}
    \caption{Performance of our baselines on the development set in the \textit{gold-two-mention} task (access to the two candidate name spans). Parallelism+URL tests the page-context setting; all other test the snippet-context setting. Bold indicates best performance in each setting.}
    \label{tab:baselines-2cands}
\end{table}

\paragraph{Structural Cues}

Baseline cues which may additionally access syntactic structure are:
\begin{itemize}
    \item \textbf{\textsc{Syntactic Distance}}. Select the syntactically closest candidate to the pronoun. Back off to \textsc{Token Distance}.
    \item \textbf{\textsc{Parallelism}}. If the pronoun is a subject or direct object, select the closest candidate with the same grammatical argument. Back off to \textsc{Syntactic Distance}.
\end{itemize}

\noindent Both cues yield strong baselines comparable to the strongest OntoNotes-trained systems (cf. Table~\ref{tab:ontonotes-baselines}).
In fact, \citet{lee2017} and \textsc{Parallelism} produce remarkably similar output:
of the 2000 example pairs in the development set, the two have completely opposing predictions (i.e. Name~A vs. Name~B) on only 325 examples.
Further, the cues are markedly gender-neutral, improving the Bias metric by 9\% in the standard task formulation and to parity in the \textit{gold-two-mention} case.
In contrast to surface cues, having the full candidate set is helpful: mention alignment via a non-indicated candidate successfully scores 69\% of \textsc{Parallelism} predictions.

\paragraph{Wikipedia Cues}

To explore the \textit{page-context} setting, we consider a Wikipedia-specific cue:

\begin{itemize}
    \item \textbf{\textsc{URL}}. Select the syntactically closest candidate which has a token overlap with the page title. Back off to \textsc{Parallelism}.
\end{itemize}

\noindent The heuristic gives a performance gain of 2\% overall compared to \textsc{Parallelism}.
That the feature is not more helpful again validates our methodology for extracting diverse examples.
We expect future work to greatly improve on this baseline by using the wealth of cues in Wikipedia articles, including page text.

\subsection{Transformer Models for Coreference}

The recent Transformer model \citep{vaswani:etal:2017} demonstrated tantalizing representations for coreference:
when trained for machine translation, some self-attention layers appear to show stronger attention weights between coreferential elements.\footnote{See Figure~4 at \url{https://arxiv.org/abs/1706.03762}}
\citet{voita2018transformers} found evidence for this claim for the English pronouns \emph{it}, \emph{you}, and \emph{I} in a movie subtitles dataset \citep{lison2018}.
GAP allows us to explore this claim on Wikipedia for ambiguous personal pronouns.
To do so, we investigate the heuristic:
\begin{itemize}
    \item \textbf{\textsc{Transformer}}. Select the candidate which attends most to the pronoun.
\end{itemize}

{
\setlength{\tabcolsep}{1pt} 
\renewcommand{\arraystretch}{0} 
\setlength{\fboxsep}{2mm} 

\begin{table}
    \centering \small
    \hspace{-1cm}
    \begin{tabular}{p{2.1cm}|RRRRRR}
    \backslashbox{Head}{Layer}
                   & \multicolumn{1}{c}{L0}   & \multicolumn{1}{c}{L1}   
                   & \multicolumn{1}{c}{L2}   & \multicolumn{1}{c}{L3} 
                   & \multicolumn{1}{c}{L4}   & \multicolumn{1}{c}{L5}   \\ 
    \hline
    \centering{H0}  & 46.9 & 47.4 & 45.8 & 46.2 & 45.8 & 45.7 \\
    \centering{H1}  & 45.3 & 46.5 & 46.4 & 46.2 & 49.4 & 46.3 \\
    \centering{H2}  & 45.8 & 46.7 & 46.3 & 46.5 & 45.7 & 45.9 \\
    \centering{H3}  & 46.0 & 46.3 & 46.8 & 46.0 & 46.6 & 48.0 \\
    \centering{H4}  & 45.7 & 46.3 & 46.5 & 47.8 & 45.1 & 47.0 \\
    \centering{H5}  & 47.0 & 46.5 & 46.5 & 45.6 & 46.2 & 52.9 \\
    \centering{H6}  & 46.7 & 45.4 & 46.4 & 45.3 & 46.9 & 47.0 \\
    \centering{H7}  & 43.8 & 46.6 & 46.4 & 55.0 & 46.4 & 46.2 \\
    \end{tabular}
    \caption{Coreference signal of a Transformer model on the validation dataset, by encoder attention layer and head.}
    \label{tab:transformer-baselines}
\end{table}
}

\noindent
The Transformer model underlying our experiments is trained for 350k steps on the 2014 English-German NMT task,\footnote{\url{http://www.statmt.org/wmt14/translation-task.html}} using the same settings as \citet{vaswani:etal:2017}.
The model processes texts as a series of \textit{subtokens} (text fragments the size of a token or smaller) and learns three multi-head attention matrices over these, two self-attention matrices (one over the subtokens of the source sentences and one
over those of the target sentences),
and a cross-attention matrix between the source and target.
Each attention matrix is decomposed into a series of feed-forward \textit{layers}, each composed of discrete \textit{heads} designed to specialize for different dimensions in the training signal.
We input GAP snippets as English source text and extract attention values from the source self-attention matrix;
the target side (German translations) is not used.

We calculate the attention between a name and pronoun to be the mean over all subtokens in these spans;
the attention between two subtokens is the sum of the raw attention values between all occurrences of those subtoken strings in the input snippet.
These two factors control for variation between Transformer models and the spreading of attention between different mentions of the same entity.

\paragraph{\textsc{Transformer-Single}} Table~\ref{tab:transformer-baselines} gives the performance of the \textsc{Transformer} heuristic over each self-attention head on the development dataset.
Consistent with the observations by \citet{vaswani:etal:2017}, we observe that the coreference signal is localized on specific heads and that these heads are in the deep layers of the network (e.g. L3H7).
During development, we saw that the specific heads which specialize for coreference are different between different models.

\begin{table}[]
    \centering
    \begin{tabular}{ll|cc}
                                      &           & \multicolumn{2}{c}{\textsc{Parallelism}}   \\
                                      &           & Correct & Incorrect \\
    \hline
    \multirow{2}{*}{\textsc{Transf.}} & Correct   &  48.7\% &   13.4\%  \\
                                      & Incorrect &  21.6\% &   16.3\%  \\
    \end{tabular}
    \caption{Comparison of the predictions of the \textsc{Parallelism} and \textsc{Transformer-Single} heuristics over the GAP development dataset.}
    \label{tab:transformer-confusion}
\end{table}

The \textsc{Transformer-Single} baseline in Table~\ref{tab:baselines} is the one set by L3H7 in Table~\ref{tab:transformer-baselines}.
Despite not having access to syntactic structure, \textsc{Transformer-Single} far outperforms all surface cues above.
That is, we find evidence for the claim that Transformer models implicitly learn language understanding relevant to coreference resolution.
Even more promising, we find that the instances of coreference that \textsc{Transformer-Single} can handle is substantially different from those of \textsc{Parallelism}, see Table~\ref{tab:transformer-confusion}.

\paragraph{\textsc{Transformer-Multi}} We learn to compose the signals from different self-attention heads using extra tree classifiers \citep{geurts2006}.\footnote{\url{http://scikit-learn.org/stable/modules/generated/sklearn.ensemble.ExtraTreesClassifier.html}}
We choose this classifier since we have little available training data and a small feature set.
Specifically, for each candidate antecedent, we:

\begin{itemize}
    \item Extract one feature for each of the 48 Transformer heads. The feature value is True if there is a substring overlap between the candidate and the prediction of \textsc{Transformer-Single}.
    \item Use the ${\chi}^2$ statistic to reduce dimensionality. We found k=3 worked well.
    \item Learn an extra trees classifier over these three features with the validation dataset.
\end{itemize}

\noindent
That \textsc{Transformer-Multi} is stronger than \textsc{Transformer-Single} in Table~\ref{tab:baselines} suggests that different self-attention heads encode different dimensions of the coreference problem. Though the gain is modest when all mentions are under consideration,  Table~\ref{tab:baselines-2cands} shows a 4.2\% overall improvement over \textsc{Transformer-Single} for the \textit{gold-two-mention} task. Future work could explore filtering the candidate list presented to Transformer models to reduce the impact of distractor mentions in a pronoun's context---for example, by gender in the \textit{page-context} setting. It is also worth stressing that these models are trained on very little data (the GAP validation set). These preliminary results suggest that learned models incorporating such features from the Transformer and using more data are worth exploring further.

\subsection{GAP Benchmarks}

\begin{table}[]
    \centering
    \begin{tabular}{l|rrrr}
                     & \multicolumn{1}{c}{M} & \multicolumn{1}{c}{F} & \multicolumn{1}{c}{B} & \multicolumn{1}{c}{O} \\
    \hline
    \citet{lee2017}  & 67.7 & 60.0 & \textit{0.89} & 64.0 \\
    Parallelism      & 69.4 & 64.4 & \textit{0.93} & 66.9\\
    \hline
    Parallelism+URL  & 72.3 & 68.8 & \textit{0.95} & 70.6 \\
    \end{tabular}
    \caption{Baselines on the GAP challenge test set.}
    \label{tab:task-baselines}
\end{table}

Table~\ref{tab:task-baselines} sets the baselines for the GAP challenge.
We include the off-the-shelf system which performed best Overall on the development set \citep{lee2017}, as well as our strongest baseline for the two task settings, \textsc{Parallelism}\footnote{We also trained an Extra Tree classifier over all explored coreference-cue baselines (including Transformer-based heuristics), but its performance was similar to \textsc{Parallelism} and the predictions matched in the vast majority of instances.} and \textsc{URL}.

\begin{table}[]
    \centering
    \begin{tabulary}{190pt}{cC|rr}
    \multirow{2}{*}{Difficulty} & Agreement & \multirow{2}{*}{\#}   & \multirow{2}{*}{\%} \\
                                & with Gold &                       & \\
    \hline
    \textit{Green} & 4 & 631 & 28.7 \\
    \hline
    \multirow{3}{*}{\textit{Yellow}} & 3 & 469 & 21.3 \\
    & 2 & 420 & 19.1 \\
    & 1 & 353 & 16.0 \\
    \hline
    \textit{Red} & 0 & 328 & 14.9 \\
    \end{tabulary}
    \caption{Analysis of the GAP development examples by the number of systems (out of 4) agreeing with gold.}
    \label{tab:agreement}
\end{table}

We note that strict comparisons cannot be made between our \textit{snippet-context} baselines given that \citet{lee2017} has access to OntoNotes annotations that we do not, and we have access to pronoun ambiguity annotations that \citet{lee2017} do not.

\begin{table*}[]
    {\small
    \begin{tabular}{p{2cm}p{4cm}p{8.5cm}r}
    \centering{Category} & Description & Example (abridged) & \# \\
    \hline
    \centering{\textsc{Narrative Roles}}
        & Inference involving the roles people take in described events
        & As \emph{Nancy} tried to pull \underline{\emph{Hind}} down by the arm in the final meters as what was clearly an attempt to drop \textbf{her} [...]
        & 28 \\
    \hline
    \centering{\textsc{Complex Syntax}}
        & Syntactic cues are present but in complex constructions
        & \emph{Sheena} thought back to the 1980s  [...] and thought of idol \underline{\emph{Hiroko Mita}}, who had appeared on many posters for medical products, acting as if \textbf{her} stomach or head hurt
        & 20 \\
    \hline
    \centering{\textsc{Topicality}}
        & Inference involving the entity topicality, inc. parentheticals
        & The disease is named after \underline{\emph{Eduard Heinrich Henoch}} (1820--1910), a German pediatrician (nephew of \emph{Moritz Heinrich Romberg}) and \textbf{his} teacher
        &  15 \\
    \hline
    \centering{\textsc{Domain Knowledge}}
        & Inference involving knowledge specific to a domain, e.g. sport
        & The half finished 4--0, after Hampton converted a penalty awarded against \emph{\underline{Arthur Knight}} for handball when \emph{Fleming}'s powerful shot struck \textbf{his} arm.
        &  6 \\
    \hline
    \centering{\textsc{Error}}
        & Annotation error, inc. truly ambiguous cases
        & When \textbf{she} gets into an altercation with \emph{Queenie}, \emph{\underline{Fiona}} makes her act as Queenie's slave [...]
        &  6 \\
    \end{tabular}
    }
    \caption{Fine-grained categorization of 75 \textit{Red} examples from the GAP development set (no system agreed with the worker-selected name). Underlining indicates the rater-selected name.}
    \label{tab:red-cases}
\end{table*}

\section{Error Analysis}

We have shown that GAP is challenging for both off-the-shelf systems and our baselines.
To assess the variance between these systems and gain a more qualitative understanding of what aspects of GAP are challenging, we use the number of off-the-shelf systems that agree with the rater-provided labels (Agreement with Gold) as a proxy for difficulty.
Table~\ref{tab:agreement} breaks down the name-pronoun examples in the development set by Agreement with Gold (the smaller the agreement the harder the example).\footnote{Given that system predictions are not independent for the two candidate names for a given snippet, we only focus on the positive coreferential name-pronoun pair when the gold label is either ``Name A'' or ``Name B''; we use both name-pronoun pairs when the gold label is "Neither".}

Agreement with Gold is low (average 2.1) and spread.
Less than 30\% of the examples are successfully solved by all systems (labeled \textit{Green}), and just under 15\% are so challenging that none of the systems gets them right (\textit{Red}).
The majority are in between (\textit{Yellow}).
Many \textit{Green} cases have syntactic cues for coreference, but we find no systematic trends within \textit{Yellow}.

Table~\ref{tab:red-cases} provides a fine-grained analysis of 75 \textit{Red} cases.
When labeling these cases, two important considerations emerged:
(1)~labels often overlap, with one example possibly fitting into multiple categories; and
(2)~GAP requires global reasoning---cues from different entity mentions work together to build a snippet's interpretation.
The \textit{Red} examples in particular exemplify the challenge of GAP, and point toward the need for multiple modeling strategies to achieve significantly higher scores on the dataset.

\section{Conclusions}

We have presented a dataset and a set of strong baselines for a new coreference task, GAP.
We designed GAP to represent the challenges posed by real-world text, in which ambiguous pronouns are important and difficult to resolve.
We highlighted gaps in the existing state of the art, and proposed the application of Transformer models to address these.
Specifically, we show how traditional linguistic features and modern sentence encoder technology are complementary.

Our work contributes to the emerging body of work on the impact of bias in machine learning.
We saw systematic differences between genders in analysis;
this is consistent with many studies which call out differences in how males and females are discussed publicly.
By rebalancing our dataset for gender, we hope to reward systems which are able to capture these complexities fairly.

It has been outside the scope of this paper to explore bias in other dimensions, to analyze coreference in other languages, and to study the impact on downstream systems of improved coreference resolution.
We look forward to future work in these directions.

\section*{Acknowledgments}

We would like to thank our anonymous reviewers and the Google AI Language team, especially Emily Pitler, for the insightful comments that contributed to this paper. Many thanks also to the Data Compute team, especially Ashwin Kakarla, Henry Jicha and Daphne Luong, for their help with the annotations, and thanks to Llion Jones for his help with the Transformer experiments.

\bibliography{main}

\end{document}